# Machine learning for Earth System Science (ESS): A survey, status and future directions for South Asia

Manmeet Singh[1,2,3]*, Bipin Kumar[1], Rajib Chattopadhyay[1], K Amarjyothi[6], Anup K Sutar[7], Sukanta Roy[7], Suryachandra A Rao[1], Ravi S. Nanjundiah[1,4,5]

[1]Indian Institute of Tropical Meteorology, Ministry of Earth Sciences, Pune, 411008, India
[2]Jackson School of Geosciences, The University of Texas at Austin, Austin, USA
[3]IDP in Climate Studies, Indian Institute of Technology Bombay, Mumbai, 400076, India
[4]Centre for Atmospheric and Oceanic Sciences, Indian Institute of Science, Bengaluru, 560012,India
[5]Divecha Centre for Climate Change, Indian Institute of Science, Bengaluru, 560012, India
[6]National Center for Medium Range WeatherForecasting, Ministry of Earth Sciences, Noida, 201309, India
[7]Borehole Geophysics Research Laboratory, Ministry of Earth Sciences, Karad, 415114, India
*email: manmeet.singh@utexas.edu

**Abstract:**
This survey focuses on the current problems in Earth systems science where machine learning algorithms can be applied. It provides an overview of previous work, ongoing work at the Ministry of Earth Sciences, Gov. of India, and future applications of ML algorithms to some significant earth science problems. We provide a comparison of previous work with this survey, a mind map of multidimensional areas related to machine learning and a Gartner's hype cycle for machine learning in Earth system science (ESS). We mainly focus on the critical components in Earth Sciences, including atmospheric, Ocean, Seismology, and biosphere, and cover AI/ML applications to statistical downscaling and forecasting problems.

**Keywords:** Machine learning, Mind map, AI/ML applications in Earth Science. Gartner hype curve

## 1. Introduction

. The recent increase in computational power has given rise to application to novel techniques. In the last few decades in conjunction with increasing computational power we notice significant improvement in forecasts at various scales using numerical techniques. The advent of satellites, modern instruments and advanced global/regional modelling capabilities has helped in amassing large amounts of data surpassing petabytes per day. Hence the need of the hour is to exploit this data innovatively. These datasets have been collected using sensors that monitor magnitudes of states, fluxes, and more intensive or time/space-integrated variables. The Earth system data exemplify all four of the "Four V's of Big Data" concerning volume, velocity, variety, and veracity. Looking at the big picture shows that our capacity to gather and store data vastly outpaces our ability to access it, much alone comprehend it meaningfully. The power to make accurate predictions has not risen in parallel with data abundance. We will need to undertake two significant endeavors to maximize the wealth of Earth system data growth and diversity, which are

(1) identifying and utilizing data insights, and

(2) developing predictive models that can discover previously unknown laws of nature while still honoring our evolving understanding of the laws of nature.

Advances in computing capacity and enhanced data availability provide exceptional new prospects. For example, machine learning and artificial intelligence technologies are now accessible, but they require additional development and adaptation to geoscientific study. In both spatial and temporal domains, new methods present new opportunities, new problems, and ethical demands for contemporary study fields in ESS[1].

Machine learning algorithms have grown with data availability, successfully applying to many geoscientific processing schemes, including the atmosphere, the land surface, and the ocean. Land cover and cloud classifications have been precursors to the GIS field since the resurgence of neural networks, thanks to the availability of very high-resolution satellite data. The majority of machine learning research in methodology (for example, kernel techniques or random forests) has since been applied to geoscience and remote sensing issues. That is frequently the case with new data appropriate for the specific approaches. In other words, machine learning has emerged as a versatile method for geoscientific data categorization. For example, change and anomaly detection issues may now be tackled using these techniques. In addition, deep learning has been employed in geoscience in the past several years to exploit better spatial and temporal patterns in the data, aspects that standard machine learning often struggles. Machine learning finds applications throughout all of ESS, and more algorithms are being incorporated into operational systems. New patterns are being discovered, in addition to the knowledge used to assess the many Earth system models.

## 1.1 Need for machine learning in ESS

Machine learning aims to uncover the transformation functions which map to the fields of high interest such as precipitation, temperature, and others. The developments in the physical sciences associated with simple statistical methodologies have left a large gray area in uncovering the relationships leading to complex, non-linear variables. There is a need to dedicate resources to using advanced machine learning based tools to decipher the links to physical fields which are still out of our reach and improve their predictability. Developments in deep learning, deep reinforcement learning, transformers, non-linear science and recent advances in interpretable machine learning are the areas that can help to solve crucial research problems in ESS. Recognizing this need to effectively utilize this large data effectively, the Ministry of Earth Sciences recently setup a virtual centre for Artificial Intelligence and Machine Learning devoted to Earth Sciences and anchored at the Indian Institute of Meteorology, Pune.

## 1.2 Related surveys

Previous surveys on the use of machine learning in ESS are summarized in Table 1. These reviews have mostly focused on the broad applications of machine learning in Earth science problems. Rolnick et al. 2019[2] focus is the most elaborated review yet on the topic, but they focused in general on the solutions to tackle the problems associated with climate change using machine learning. Others focused more on hydrology or remote sensing problems, with Reichstein et al. 2019[3] being the nearest survey to the one we have worked upon in this paper.

Table 1: Comprehensive summary of the previous surveys on machine learning in ESS and the comparison with this survey

| Previous reviews | A | B | C | D | E | F | G | H | I | J | K | L | M | N | O | P | Q | R | S | T | U | V | W | X | Y | Z |
|---|---|---|---|---|---|---|---|---|---|---|---|---|---|---|---|---|---|---|---|---|---|---|---|---|---|---|
| Rolnick et al. 2019[2] | √ | √ | √ | √ | √ | √ | √ | √ | √ | √ | √ | √ | √ | √ | √ | √ | √ | | | | | | | | | |
| Reichstein et al. 2019[3] | | | | | | | √ | | √ | √ | √ | | | √ | | √ | √ | | | | | √ | √ | √ | | √ |
| Shen et al. 2018[4] | | | | | | | √ | | √ | √ | √ | | | √ | √ | √ | √ | | √ | | | √ | √ | | | √ |
| Sit et al. 2020[5] | | | √ | √ | √ | | √ | | | √ | | √ | √ | √ | √ | √ | √ | | | | | | √ | √ | √ | |
| Ball et al. 2017[6] | | √ | √ | | √ | | √ | | | √ | | | | √ | √ | √ | √ | | √ | | | | √ | √ | √ | |
| Fang et al. 2021[7] | √ | √ | √ | | | √ | | √ | √ | | √ | | √ | √ | | | | | √ | | | √ | | | | √ |
| This review | √ | √ | √ | √ | √ | √ | √ | √ | √ | √ | √ | √ | √ | √ | √ | √ | √ | √ | √ | √ | √ | √ | √ | √ | √ | √ |

*Abbreviations:* A-Electricity Systems, B- Transportation systems, C- Buildings & Cities/Urban climate, D- Industrial systems, E- Farms & Forests, F- Climate change mitigation, G- Weather & Climate prediction, H- Climate finance, I- Causality, J- Computer vision, K-Interpretable machine learning, L-Natural language processing, M-Reinforcement learning, N-Time series, O-Transfer learning, P-Uncertainty estimation, Q-Unsupervised learning, R- Seismology, S- South Asian Monsoon, T- Short-range weather prediction, U-Extended range weather forecasting, V- Seasonal weather prediction, W- Hydrology, X-Oceanography, Y- Transformers or Generative adversarial networks, Z- weather and climate extremes

## 1.3 Motivation for this study

- The previous surveys have only addressed the problems within ESS in general. There is a need for a review paper focusing on the studies and issues addressing the South Asian region. For example, the Indian monsoon is one of the most complex climate phenomena whose mystery is yet not fully solved. It requires special focus and attention to address the challenges in accurately predicting the various spatiotemporal scales of the monsoon.
- The studies summarized in Table 1 have not considered the latest state-of-the-art algorithms such as the attention-based Transformers and the Generative Adversarial Networks. The advancements brought by these models in the computer vision and natural language processing community make them excellent candidates to be explored in the domain of ESS.
- This review outlines all the previous review papers on the subject, delineates the tools required, the material required by anyone to gain hands-on experience in machine learning and can be used to further the applications of machine learning in ESS.

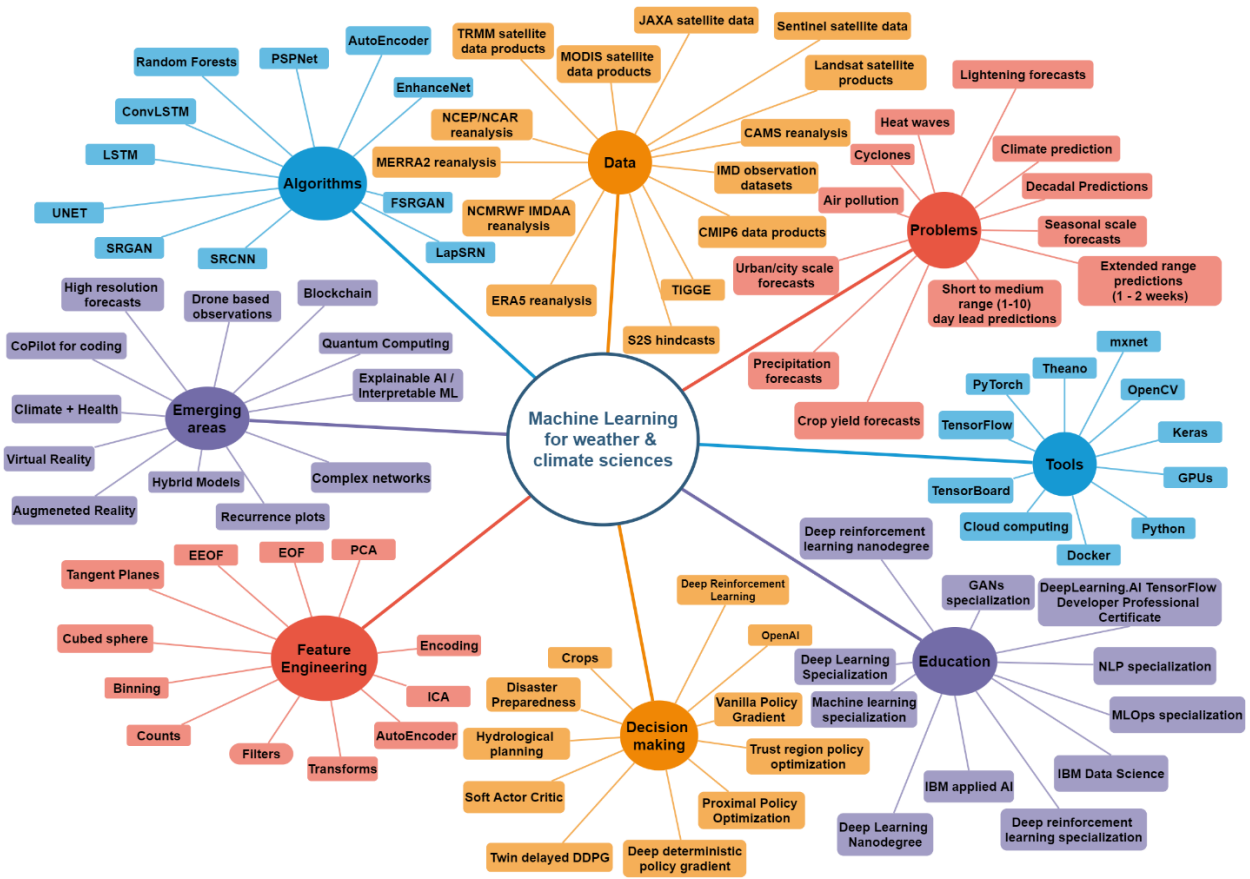

Figure 1: Mind map of multidimensional areas related to machine learning in weather and climate sciences

## 2. Background

This section discusses the algorithms, data, problems, tools, educational material, feature engineering, and the emerging areas related to machine learning in Earth sciences. These have been summarized in the mind map, depicted in Figure 1, taking the case of weather and climate sciences as an example.

### 2.1 Machine learning algorithms for ESS

Various algorithms which have shown remarkable performance in computer vision, natural language processing, and reinforcement learning etc can be directly applied to the problems of ESS. For example, the super-resolution methodology (SRCNN, DeepSD) developed by Dong et al. (2015)[8] to enhance the resolution of image datasets has been used to downscale the precipitation datasets from coarser resolution to a high resolution[9,10]. Seasonable forecast of various aspects of monsoon has been studied using single and stacked encoder-based techniques[11,12]. Prediction of solar irradiance using CNN with added attention has been used[12]. Recent advances in the computer vision community show that algorithms such as SRGAN, LapSRN, FSRGAN, and UNET outperform the standard SRCNN. Long short-term memory (LSTM) networks, sequence to sequence networks, and the recent attention-based Transformer models have improved the accuracies in natural language processing. Some of these algorithms have also been used or can be applied to the time series forecasting problems in ESS. For example a survey on these applications can be found in Lim and co-authors[13] . Weather and climate data are so massive that it is not even encountered by the community working on big data. The spatio-temporal nature of the datasets, i.e., three-dimensional fields at each temporal dimension, makes it a complex problem to solve. The patterns in this four-dimensional data cannot be deciphered manually, and machine learning offers the perfect opportunity. Models that have shown good performance on video datasets such as ConvLSTM can be used to build large-scale deep learning-based systems which can predict the information in high spatial and temporal resolutions[14,15]. Sequence-to-sequence and LSTM has been used to predict and forecast active-break cycles of Indian monsoon[16] . Before starting any analysis, traditional algorithms such as random forests, support vector machines and multivariate linear regression should be the first goto methods. EnhanceNeT and PSPNet are algorithms which can be used for classification of objects in images and spatially locating them. They have shown excellent results in computer vision applications. They can be used for problems such as the identification of floods from satellite imagery.

## 2.2 ESS datasets

The understanding of ESS datasets is most important while developing machine learning models. These datasets primarily come in three classes, viz the observational data, reanalysis product combination of model data and observation created in a consistent manner) , dynamical model simulated outputs (such as climate change data from models) . For the South Asian domain, long-period ground-based observations from the India Meteorological Department (IMD) are available. These datasets can now be obtained from https://dsp.imdpune.gov.in. Satellite- based products are available from the Tropical Rainfall Measuring Mission (TRMM), Landsat, Sentinel and MODIS. Reanalysis products are gridded products and are very useful for the fields which are/cannot be measure directly by instruments. They offer insights into the information which is nearest to reality. Various reanalysis products are available for the South Asian region, such as

- (i) IMDAA reanalysis,
- (ii) NCEP/NCAR reanalysis,
- (iii) CAMS reanalysis,
- (iv) ERA5 reanalysis,
- (v) MERRA-2 reanalysis and
- (vi) JRA55 reanalysis.

As regards model products are considered, various model products such as TIGGE, CMIP5/CMIP6, and Seasonal to sub-seasonal (S2S) hindcasts are available. The model outputs are based on the integrations of partial differential equations based on dynamical systems. Machine learning offers an innovative methodology to improve these dynamical model estimates by combining them with observed or reanalysis products. The archive of seismic waveform data, Global Positioning System (GPS) data, oceanographic and other geoscience datasets in the country are increasing exponentially every year, calling for fast and efficient processing and dissemination of information to the public service systems.

## 2.3 Research problems in ESS

South Asia is home to more than two billion inhabitants and is heavily dependent on the natural climate variability for livelihood. For example, the Indian monsoon feeds agricultural lands over the region, thus directly impacting the region's economic well-being. Monsoon is a complex multi-scale problem and is nonlinear and hence linear methods cannot unravel the real processes especially the feedback processes leading to its variability. Forecasts at various temporal scales such as short to medium range (1-10 days), extended range (2-3 weeks), seasonal scale (for coming season) and climate scale (100s year) are important for planning hydrological resources over the region. It has been known that the crop yields are dependent on the meteorological variables; machine learning can be used to accurately forecast the spatial crop yield a season in advance and economically benefit the society. The demographics in the South Asian region have considerably changed in the past decades and many people now live in the cities.

Moreover, the population density in these areas is very high. Hence, locally accurate urban forecasts are need of the hour. These locations are also sources of chemical species which are harmful for the environment and all living beings. Hence air pollution prediction is an important problem. Identifying localities with higher air pollution is also essential for city planning, for example, deciding the number of electric buses by the city authorities. Machine learning-based algorithms can be used to improve the cyclone forecasts of dynamical models. Extreme weather events such as heat waves and cloud bursts are causing havoc in recent times, as it is challenging to predict them accurately. In seismology, AI/ML techniques are being tried out for earthquake detection, phase-picking (measurement of arrival times of distinct seismic phases), event classification, earthquake early warning, ground motion prediction, tomography and earthquake geodesy.

**2.4 Popular tools to perform machine learning for ESS**

The availability of open-source software packages has provided a bridge to the domain experts to avoid reinventing the wheel while applying machine learning to their problems. Python is the most popular language for machine learning, and various libraries such as TensorFlow, PyTorch, Theano, mxnet, OpenCV, Keras, and others are available freely. Visualization software such as TensorBoard and Tableau assist in communicating the results from machine learning models. In addition to the software requirements, deep learning needs Graphical Processing Units (GPUs) to perform the tensor computations in neural networks. Tensor Processing Units (TPUs) are a step ahead of GPUs wherein the neural network is encoded on the chip to perform fast calculations. The TPUs are only available over the cloud, and every individual can't buy a personal GPU for deep learning. Hence, free and paid cloud computing services, such as Amazon Web Services (AWS), Microsoft Azure, Google Cloud Platform (GCP), Paperspace, Digital Ocean, Google Earth Engine provide an option to build machines over the cloud to perform deep learning and conduct analysis in climate science[17]. A step further, the concept of Jupyter notebooks as a service has become popular and there are several frees as well as paid vendors providing notebook as a service. Notable amongst them include the free services provided by Kaggle, Google Colab and others. More cloud vendors can be found at https://github.com/binga/cloud-gpus, https://github.com/zszazi/Deep-learning-in-cloud, https://github.com/discdiver/deep-learning-

cloud-providers/blob/master/list.md and others actively maintained. "Docker containers" have also become an important part of the ecosystem helping deploy end-to-end packages for deep learning.

## 2.5 Educational materials to learn machine learning for ESS

A key component in the machine learning cycle is the educational resources to build knowledge and apply it to Earth and climate science. The avenues to learn data science and hence use machine learning for Earth science applications are the Coursera specializations, courses, professional certificates, Udacity nanodegrees, Udemy courses and other free and paid material available as MOOCs. The Development of Skilled Manpower in Earth System Sciences (DESK) at the Ministry of Earth Sciences (MoES), Government of India regularly holds training programs to train young researchers on machine learning applications in in Earth sciences. One such training workshop was conducted in 2021 and the details can be found at https://www.youtube.com/watch?v=2cQmhDgFinQ&list=PLgQCKqNw6z_CMKiKoWMAukvo8vyEe1KLb

## 2.6 Decision making for machine learning in ESS

After the weather/hydrological forecasts have been generated, they have to be used to take decisions for the benefit of society. Deep reinforcement learning is an excellent methodology for this purpose. State of the art algorithms such as deep q networks, vanilla policy gradient, trust region policy optimization, proximal policy optimization, deep deterministic policy gradient (DDPG), soft actor critic, twin delayed DDPG and others can be used to train agents which can guide in decision making. The most crucial aspect of deep reinforcement learning is the design of the environment, action(s), and reward(s). The decision-making can be used for disaster preparedness/mitigation, hydrological planning, and other associated tasks.

## 2.7 Feature engineering for machine learning in ESS

Feature engineering is the generation of meaningful predictors or parameters to improve the performance of a machine learning model. It is performed after cleaning the data and preparing it in a format that can train statistical models. It has been noted that removing redundant variables improves the performance of machine learning systems. Various methods can be used to find the most valuable predictors some of them are: Principal Component Analysis (PCA), Empirical Orthogonal Functions (EOF), and Independent Component Analysis (ICA). Binning, counts, transforms or filtering can be used to extract the predictive signal from data to improve the models. Unsupervised learning techniques such as autoencoder can also assists in finding valuable predictors from raw datasets. The deep learning-based models are, however, coded for image-based input datasets. To overcome this limitation, strategies such as transforming the spherical global data to a cubed sphere or tangent planes mapping can effectively reduce spherical distortions in the data.

## 2.8 Emerging areas in machine learning for ESS

While the previous decade has seen the hype of deep learning overshadow other machine learning methodologies, there are numerous emerging and innovative machine learning methods which can be used for ESS. Graph machine learning is training neural networks on graphs and is becoming increasingly popular. Complex networks and recurrence plots come in the category of nonlinear methodologies and come in handy for specific applications. One major concern while using machine learning for the physical sciences is that these models are known as black box models. Interpretable machine learning aims to address this concern and analysis of deep learning model weights reveal the patterns learned. Ongoing active research is happening in this area, and it is crucial for the increasing acceptability of deep learning models at the production scale in ESS. The emerging fields of augmented reality, virtual reality, improved remote sensing measurements, crowd-sourcing and drone technology offer excellent potential to advance the observation data collection and improve machine learning models.

## 3. Applications of artificial intelligence and machine learning in Earth sciences

The A/ML algorithms have vast applications in the Earth Science problems. Figure 2 depicts few applications in the areas, including atmosphere/biosphere, seismology, and ocean.

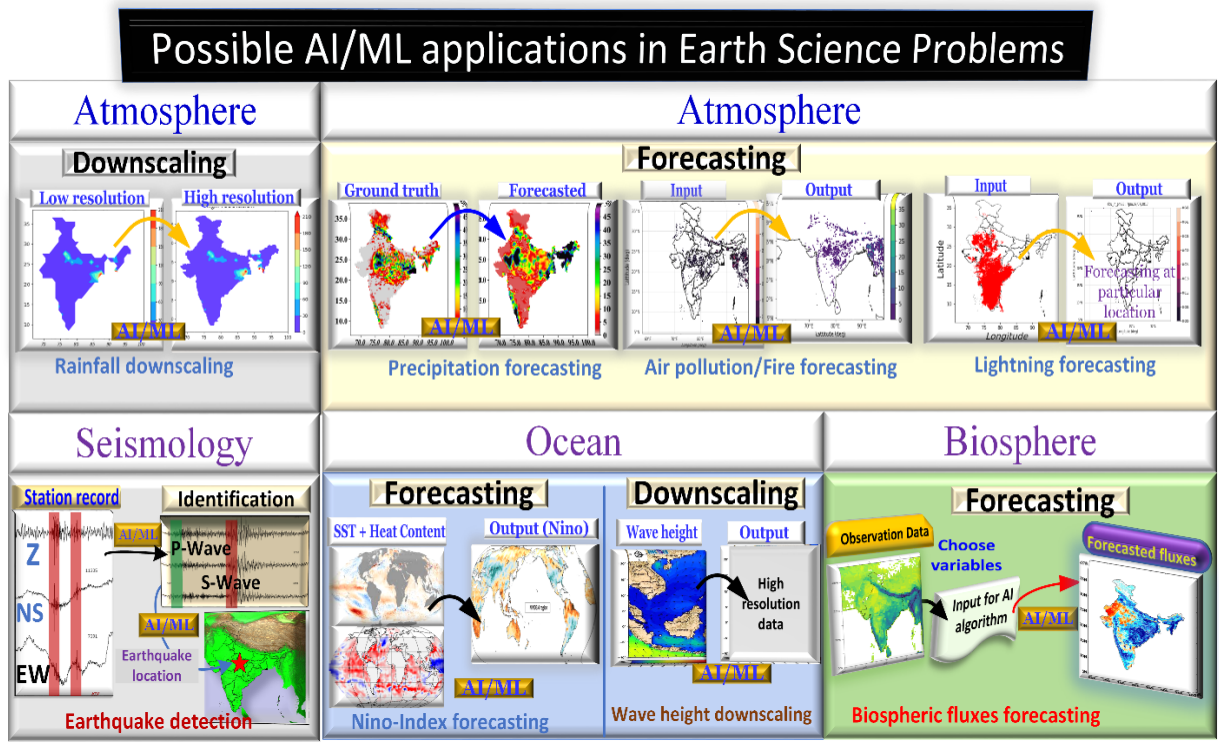

Figure 2: An overview of the application of AI/ML algorithms in few areas from Earth Science problems. The precipitation forecasting can include data from short-range, medium-range and extended-range forecasting.

## 3.1 Statistical downscaling

Downscaling of data is necessary to get a local projection of the information. The present-day models and observations generated from weather stations (or other instruments) are available at coarser resolution and are irregularly spaced, which may often provide misrepresentation (or absence) of precipitation, temperature, or other variables at local levels. Downscaling of Indian Summer Monsoon (ISM) rainfall is a difficult task as it involves a multi-scale spatiotemporal dynamical process with significant variance[18]. Further, regional variations of ISM rainfall are often quite large varying from a few millimeters to thousands of millimeters within a few hundred kilometers. The ISM rainfall can be classified into different coherently fluctuating zones, which may be linked to complex multi-scale processes[19–21].

The statistical downscaling is a low-cost method to obtain information at the local scale and provide it to stockholders. Artificial Intelligence (AI) and Machine Learning (ML) techniques are being used for statistical downscaling[8,22]. Recently, development in the Single Image Super-Resolution (SR) using Deep Learning proved to be one of the best methods used for this purpose[8–10]. Another method that showed promising results in the statistical downscaling is the ConvLSTM method documented by Harilal and coauthors[23].

## 3.2 Seismological events

The growing volume of seismological and other geoscience-related datasets acquired from surface and borehole studies, require efficient techniques for analysis and trend recognition to extract valuable signals. AI/ML tools have been applied in different fields in seismology, from event identification to earthquake prediction, with varying degrees of success[24–29] The case studies also bring out the need for further research and development to refine the existing techniques and develop new tools that could be utilised in the processing and analyses of large datasets and identification of different geophysical signals. The use of AI/ML techniques in geoscience / seismological could be employed gainfully to analyse other seismological datasets that are routinely acquired by MoES and its affiliated institutions. Identifying seismic phases, accurately, is one of the primary requirements for seismological data analysis towards the determination of earthquake source parameters. ML is planned to be used in identifying different seismic phases in the data.

In many earthquake detection algorithms, STA/LTA criteria are being used to detect possible arrival times of P and S waves[30]. Later, matched filtering or template matching technique was used for event detection. In this method, waveforms of known events are used as templates to scan through continuous waveforms to detect new events[31]. Recently, ML is utilized to improve earthquake detection and phase picking capabilities[25,32]. Fingerprinting and Similarity Thresholding (FAST) is the latest algorithm using ML techniques to identify earthquakes without prior knowledge of seismicity. Being computationally more efficient than template matching FAST would facilitate automated processing of large voluminous datasets. Similarly, the generalized phase detection (GPD) algorithm searches for near identical waveforms from millions of seismograms, which is used to classify windowed data as P, S or noise. GPD can be applied to datasets that were not just encompassed by training sets but can also be applied to difficult cases such as clipped seismograms. Kong et al., (2018)[33] used Neural Networks (NN) for training to pick P-wave onset and to detect P-wave polarity. ML techniques have important applications in the detection of small magnitude local earthquakes in areas which are characterized by sparsity of receivers. ML/AI algorithms may play an important role in the identification of events and in locating earthquakes with recordings of the events at fewer stations[33]. Other applications in earth sciences such as hydrology, AI/ML can be used for estimating and predicting stream flow in ungauged basins[35–37].

### 3.3 Short and medium range forecasting using machine learning

Currently, the global highest resolution ensemble predication system at ~12.5 km horizontal resolution (with 21 members) is being used for providing 10 days probabilistic forecast based on the Global Ensemble Forecast System (GEFS@T1534) India Meteorological Department. The high resolution GEFS has been implemented by IITM for operational application since June 2018. While the deterministic GFS model[38] at 12.5 km provides a better skill up to ~5 days compared to the earlier coarser resolution (~25 km resolution GFST574)[39], the ensemble prediction system has shown much better skill than the control member (the deterministic GFS model) particularly for predicting extreme rainfall events[40-41]. The model forecast inaccuracies mainly arise from initial conditions and improper physical parameterizations. The uncertainties of initial conditions are largely resolved by the perturbed initial conditions in the ensemble prediction system, however, the uncertainty arising from deterministic closures of the physical parameterization still adds much errors due to unrealistic constraints, namely the quasi-equilibrium[42]. Under the AI/ML/Dl paradigm, the use of sub-grid scale tendencies generated by the cloud resolving models within each climate model grid as the input of a deep learning model for training to target the heat and moisture trained tendencies hold promise in improving the model fidelity[43–45].

### 3.4 Machine learning for extended range forecasts

AI/ML methods recently find applications in the climate forecast models. Two basic applications show promise for near future applications. The first one is the bias correction and improvement of numerical model forecast. The other one is the methods attempting the sub seasonal low frequency forecast. The bias correction and model post processing applications are of much use to the stakeholders using climate forecast. The climate forecasts from the dynamical models show a lot of bias when forecast is considered over scales lower than the balanced flow, mainly arising due to unknown physics or unresolved dynamics. In situations where enough observation are available over a location, some of the systematic errors arising due to unresolved scale dynamics or physics can be corrected[46]. Subseasonal forecasting using machine learning methods are recently under active research[12,47–49]

### 3.5 Machine learning for seasonal and climate scale forecasting

Seasonal forecasting is one of the tough problems in forecasting. As pointed out by Lorenz (1963)[50] the weather forecasts are strongly dependent on initial conditions (today's weather determines tomorrow's weather) and in contrast climate projections/decadal predictions (an average of weather for a few decades) do not suffer from the initial conditions, however, depends on boundary conditions. When we try to make seasonal forecasts, the distinction is rather blurred and the seasonal forecasts still depend on intial conditions[51]. Chattopadhyay et al (2016)[51] have shown that models hindcasts initialized with February initial conditions exhibit better prediction skill for Indian Summer Monsoon rainfall (ISMR). Further complexities such as resolving ocean processes also become important at seasonal scale. Hence, extracting predictive information (which changes from event to event) across both space and time scales is very important to make significant progress in seasonal forecasts[52]. Therefore, use of AI/ML methods for making improved seasonal forecasts is imperative and research community started using these methods widely in seasonal forecasts[53–55] and some researchers even believe that AI/ML methods can outperform conventional prediction systems[54-55].

One of the long-standing seasonal prediction problems is prediction of Indian summer monsoon rainfall. H.F. Blandford started seasonal forecasting of Indian summer monsoon using empirical methods in 1886. Since then, numerous attempts were made to predict seasonal mean monsoon over India using both empirical models and dynamical models (Atmosphere and coupled ocean-atmosphere models, see review article of Rao et al., (2019)[39] for more details). Empirical models showed very high skills (>0.9) during development stages and during actual operational phase they showed weak skills (<0.5). On the other hand, dynamical models showed moderate skill during hindcast period as well as during operational forecast[39]. The major reason for empirical models' failure to provide high skills during operational phase is that the relationship between predictor and predictands undergo secular changes from the time the model was developed to the phase when it is made operational. To avoid such a situation AI/ML models can be used efficiently to identify new predictors[53]. Using autoencoders Saha et al (2021)[53] have developed an AI/ML model to predict IMSR with two months lead time and absolute mean error less than 3%. On the other hand, the dynamical models exhibit systematic biases in precipitation (See Rao et al., 2019[39]) and basically arise due to parametrization schemes used in these models and therefore underestimate extremes. To avoid such systematic problems AI/ML models will become handy.

### 3.6 Machine learning for improving the physical processes in dynamical models

Dynamical models work on the principle of solving partial differential equations over the area of interest with the necessary initial and boundary conditions. They consist of various components such as atmosphere, ocean, land surface and others. The correct representation of physical processes in the numerical models is highly essential for accurate simulations of the coupled climate system. For example, various researchers have tried to understand the relationship of global and regional teleconnections such as ENSO[56-57], IOD[58], North Atlantic Oscillation[59], Pacific Decadal Oscillation[60], volcanic eruptions[61], aerosols[62-63] to Indian monsoon. Recent studies have attempted the use of deep learning to develop models which better represent the physical processes. For example, Witt et al (2019)[64] used deep reinforcement learning based approach to test the stratospheric aerosol injection on climate. Volcanic eruptions have been used as an analog for the stratospheric aerosol injection and deep learning can assist in addressing the non-linear nature of the problem. Recently Lamb et al (2021)[43] used graph neural networks to learn the aerosol optical properties. Similarly, Seifert et al (2020)[65] discuss the role of machine learning in estimating the cloud microphysics. The uncertainties in the simulation of Indian monsoon arise from the missing or erroneous physics in the dynamical systems. Machine learning to improve the physical processes can lead to cascading returns by improving the hydrological outputs from the numerical weather prediction models[66–70].

### 3.7 Machine learning for Nowcasting and tracking the storms cells

The need for a high-resolution early warning system with reliable nowcasts in the regions of steep topography and urban areas during severe weather is highly essential. Traditionally, Nowcasting is served by carrying out extrapolation, probabilistic Nowcasting[71] , semi-Lagrangian advection scheme[72] , and using algorithms like optical flow etc. The latest data-driven approach is playing a key role in the area of Nowcasting too. Doppler Weather Radar provides extremely high geographical and temporal resolution weather information. Agarwal et al. (2019)[73] utilised radar pictures to forecast the weather using the U-Net algorithm, demonstrating that it outperformed the optical flow technique. Su et al., (2020)[74] have shown that machine learning approaches have a high learning capacity and enhance echo position and intensity forecast accuracy in convective cells. The temporal precision of such convective cells varies from 30 to 60 minutes during a relatively short period of time. The underestimating of precipitation in complicated orography regions is a well-known problem in precipitation estimation. Arulraj and Ana, (2021)[75] used detection and classification machine learning algorithms to improve orographic precipitation across the Southern Appalachian Mountains. Machine Learning for NWP.

Satellite remote sensing and NWP groups are ripe for rapid advancement in their application of machine learning. The NWP relies heavily on the integration of fields generated by satellites and other remote sensing devices. Gaps, both spatially and temporally, are a common occurrence in such data. The existence of gaps, both spatially and temporally, is a typical issue in such observations. The time series of satellite ocean fields are constructed using an ensemble of NNs with varying weights[76], and a deep learning method to reconstruct the optical images[77]. When modeling, deploying systems, and even issuing warnings, the ML method can give a post-forecast correction to account for uncertainties after learning from all previous failures[78].

## 4. Summary and future directions

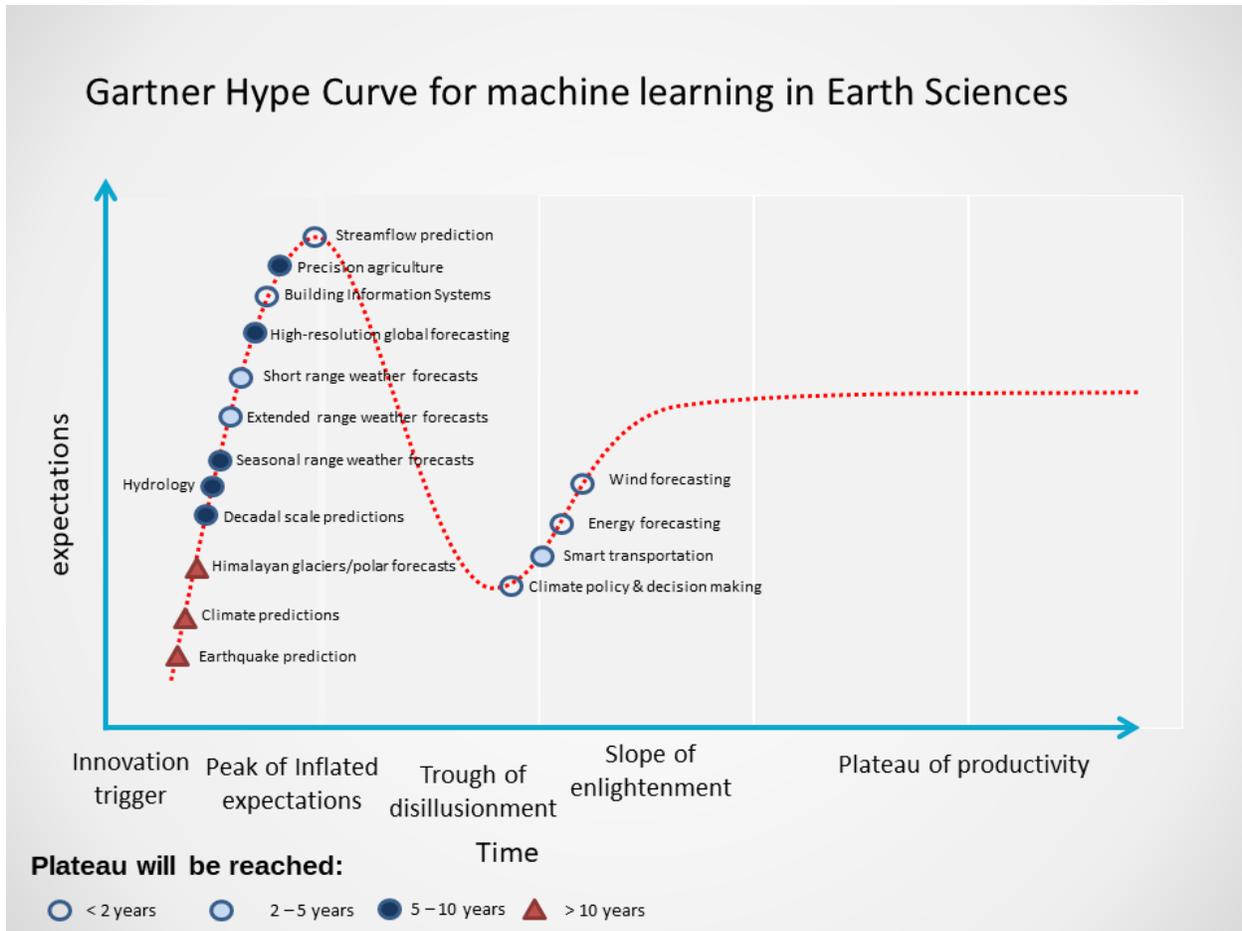

Figure 3: Gartner's hype cycle for machine learning in ESS with a focus on research problems associated with South Asia

In this study, a review of the applications of machine learning in ESS has been made. The future directions focused on solutions for the South Asian region have been summarized as a Gartner's curve in Figure 3. Hard AI problems such as that of earthquake prediction and climate scale predictions. These problems require long lead times of several years to centuries and will take more than a decade of development to be fully solved by machine learning and allied techniques. Such a long development time is expected because of the sparsity of data, for example, over the Himalayan region, under the solid Earth for earthquake prediction. The large uncertainties in dynamical models to project end of century estimates of the climate are also expected to be resolved after extensive research and development. Recent developments in machine learning, particularly in deep learning are expected to lead to transformative improvements in the short to extended range forecast, smart transportation, precision agriculture, policymaking, wind and energy forecasts during this decade. These advancements would be driven by the critical nature of these problems and the availability of high spatiotemporal drones, ground-based observations and satellite datasets.

We have discussed various AI/ML techniques used and the ones that have high potential for improving the state-of-the-art in the ESS. An exhaustive literature survey on the applications over South Asian domain, a mind map incorporating all the essential components of data science applications in ESS, and a Gartner's curve for future directions are the main contributions of this study. It can be used as a starting point to understand the existing research problems, applicable algorithms, educational resources required, hardware/software needs and other important aspects essential to work on the applications. As an end goal, this work aims to further the ESS over South Asia using machine learning applications.

**References:**

1. Chantry, M., Christensen, H., Dueben, P. & Palmer, T. Opportunities and challenges for machine learning in weather and climate modelling: hard, medium and soft AI. *Philosophical Transactions of the Royal Society A: Mathematical, Physical and Engineering Sciences* **379**, 20200083 (2021).
2. Rolnick, D. *et al.* Tackling Climate Change with Machine Learning. *arXiv.org* (2019) doi:https://arxiv.org/pdf/1906.05433.pdf.
3. Reichstein, M. *et al.* Deep learning and process understanding for data-driven Earth system science. *Nature* **566**, 195–204 (2019).
4. Shen, C. A Transdisciplinary Review of Deep Learning Research and Its Relevance for Water Resources Scientists. *Water Resources Research* **54**, 8558–8593 (2018).
5. Sit, M. *et al.* A comprehensive review of deep learning applications in hydrology and water resources. *Water Science and Technology* **82**, 2635–2670 (2020).
6. Ball, J. E., Anderson, D. T. & Chan, C. S. A Comprehensive Survey of Deep Learning in Remote Sensing: Theories, Tools and Challenges for the Community. *Journal of Applied Remote Sensing* **11**, 042609 (2017).
7. Fang, W., Xue, Q., Shen, L. & Sheng, V. S. Survey on the Application of Deep Learning in Extreme Weather Prediction. *Atmosphere* **12**, (2021).
8. Dong, C., Chen, Loy, C. C., He, K. & Tang, X. Image Super-Resolution Using Deep Convolutional Networks. *CoRR* **abs/1501.00092**, (2015).
9. Kumar, B. *et al.* Deep learning–based downscaling of summer monsoon rainfall data over Indian region. *Theoretical and Applied Climatology* **143**, 1145–1156 (2021).
10. Vandal, T. *et al.* DeepSD: Generating High Resolution Climate Change Projections through Single Image Super-Resolution. *arXiv.org* 1–9 (2017) doi:https://arxiv.org/abs/1703.03126.
11. Saha, M., Mitra, P. & Nanjundiah, R. S. Autoencoder-based identification of predictors of Indian monsoon. *Meteorology and Atmospheric Physics* **128**, 613–628 (2016).
12. Saha, M. & Nanjundiah, R. S. Prediction of the ENSO and EQUINOO indices during June–September using a deep learning method. *Meteorological Applications* **27**, e1826 (2020).
13. Lim, B. & Zohren, S. Time-series forecasting with deep learning: a survey. *Philosophical Transactions of the Royal Society A: Mathematical, Physical and Engineering Sciences* **379**, 20200209 (2021).
14. Kumar, B. *et al.* Deep Learning Based Forecasting of Indian Summer Monsoon Rainfall. **1**, (2021).
15. Shi, X. *et al.* Convolutional LSTM Network: A Machine Learning Approach for Precipitation Nowcasting. *arXiv.org* **1506.04214**, (2015).
16. Viswanath, S., Saha, M., Mitra, P. & Nanjundiah, R. S. Deep Learning Based LSTM and SeqToSeq Models to Detect Monsoon Spells of India. in *Computational Science – ICCS 2019* (2019).
17. Singh, M., Singh, B.B., Singh, R., Upendra, B., Kaur, R., Gill, S.S. and Biswas, M.S.. Quantifying COVID-19 enforced global changes in atmospheric pollutants using cloud computing based remote sensing. *Remote Sensing Applications: Society and Environment*, **22**, p.100489 (2021).
18. Chang, C.-P. *et al.* The Multiscale Global Monsoon System: Research and Prediction Challenges in Weather and Climate. *Bulletin of the American Meteorological Society* **99**, ES149–ES153 (2018).
19. Gadgil, S., Yadumani & Joshi, N. V. Coherent rainfall zones of the Indian region. *Royal Meteorologicla Society* **13**, 546–566 (1993).
20. Gadgil, S. The Indian Monsoon and Its Variability. **31**, 429–467 (2003).
21. Moron, V., Robertson, A. W. & Pai, D. S. On the spatial coherence of sub-seasonal to seasonal Indian rainfall anomalies. *Climate Dynamics* **49**, 3403–3423 (2017).
22. Tripathi, S., Srinivas, V. V. & Nanjundiah, R. S. Downscaling of precipitation for climate change scenarios: A support vector machine approach. *Journal of Hydrology* **330**, 621–640 (2006).
23. N. Harilal, M. Singh, & U. Bhatia. Augmented Convolutional LSTMs for Generation of High-Resolution Climate Change Projections. *IEEE Access* **9**, 25208–25218 (2021).

24. Bergen Karianne J., Johnson Paul A., de Hoop Maarten V., & Beroza Gregory C. Machine learning for data-driven discovery in solid Earth geoscience. *Science* **363**, eaau0323 (2019).
25. Perol Thibaut, Gharbi Michaël, & Denolle Marine. Convolutional neural network for earthquake detection and location. *Science Advances* **4**, e1700578.
26. Rouet-Leduc, B., Hulbert, C. & Johnson, P. A. Continuous chatter of the Cascadia subduction zone revealed by machine learning. *Nature Geoscience* **12**, 75–79 (2019).
27. Reynen, A. & Audet, P. Supervised machine learning on a network scale: application to seismic event classification and detection. *Geophysical Journal International* **210**, 1394–1409 (2017).
28. Kong Qingkai, Allen Richard M., Schreier Louis, & Kwon Young-Woo. MyShake: A smartphone seismic network for earthquake early warning and beyond. *Science Advances* **2**, e1501055.
29. REDDY, R. & NAIR, R. R. The efficacy of support vector machines (SVM) in robust determination of earthquake early warning magnitudes in central Japan. *Journal of Earth System Science* **122**, 1423–1434 (2013).
30. Allen, R. V. Automatic earthquake recognition and timing from single traces. *Bulletin of the Seismological Society of America* **68**, 1521–1532 (1978).
31. Gibbons, S. J. & Ringdal, F. The detection of low magnitude seismic events using array-based waveform correlation. *Geophysical Journal International* **165**, 149–166 (2006).
32. Wiszniowski, J., Plesiewicz, B. M. & Trojanowski, J. Application of real time recurrent neural network for detection of small natural earthquakes in Poland. *Acta Geophysica* **62**, 469–485 (2014).
33. Kong, Q. *et al.* Machine Learning in Seismology: Turning Data into Insights. *Seismological Research Letters* **90**, 3–14 (2018).
34. Zhu, L. *et al.* Deep learning for seismic phase detection and picking in the aftershock zone of 2008 Mw7.9 Wenchuan Earthquake. *Physics of the Earth and Planetary Interiors* **293**, 106261 (2019).
35. Besaw, L. E., Rizzo, D. M., Bierman, P. R. & Hackett, W. R. Advances in ungauged streamflow prediction using artificial neural networks. *Journal of Hydrology* **386**, 27–37 (2010).
36. Mudashiru, R. B., Sabtu, N., Abustan, I. & Balogun, W. Flood hazard mapping methods: A review. *Journal of Hydrology* **603**, 126846 (2021).
37. Zhang, D. *et al.* Intensification of hydrological drought due to human activity in the middle reaches of the Yangtze River, China. *Science of The Total Environment* **637–638**, 1432–1442 (2018).
38. Mukhopadhyay, P. *et al.* Performance of a very high-resolution global forecast system model (GFS T1534) at 12.5 km over the Indian region during the 2016–2017 monsoon seasons. *Journal of Earth System Science* **128**, 155 (2019).
39. Rao, S. A. *et al.* Monsoon Mission: A Targeted Activity to Improve Monsoon Prediction across Scales. *Bulletin of the American Meteorological Society* **100**, 2509–2532 (2019).
40. Deshpande, N. R. & Kulkarni, J. R. Spatio-temporal variability in the stratiform/convective rainfall contribution to the summer monsoon rainfall in India. *International Journal of Climatology* **n/a**, (2021).
41. Mukhopadhyay, P. *et al.* Unraveling the Mechanism of Extreme (More than 30 Sigma) Precipitation during August 2018 and 2019 over Kerala, India. *Weather and Forecasting* **36**, 1253–1273 (2021).
42. Tirkey, S., Mukhopadhyay, P., Krishna, R. P., Dhakate, A. & Salunke, K. Simulations of Monsoon Intraseasonal Oscillation Using Climate Forecast System Version 2: Insight for Horizontal Resolution and Moist Processes Parameterization. *Atmosphere* **10**, (2019).
43. Lamb, K. D. & Gentine. Zero-Shot Learning of Aerosol Optical Properties with Graph Neural Networks. (2021) doi:arXiv:2107.10197.
44. Rasp, S., Pritchard, M. S. & Gentine, P. Deep learning to represent subgrid processes in climate models. *Proc Natl Acad Sci USA* **115**, 9684 (2018).
45. Brajard, J., Carrassi, A., Bocquet, M. & Bertino, L. Combining data assimilation and machine learning to infer unresolved scale parametrization. *Philosophical Transactions of the Royal Society A: Mathematical, Physical and Engineering Sciences* **379**, 20200086 (2021).
46. Chattopadhyay, R., Sahai, A. K. & Goswami, B. N. Objective Identification of Nonlinear Convectively Coupled Phases of Monsoon Intraseasonal Oscillation: Implications for Prediction. *Journal of the Atmospheric Sciences* **65**, 1549–1569 (2008).
47. Martin, Z., Barnes, E. & Maloney, E. Predicting the MJO using interpretable machine-learning models. *Earth and Space Science Open Atchive* (2021) doi:https://doi.org/10.1002/essoar.10506356.1.
48. Borah, N., Sahai, A. K., Chattopadhyay, R., Joseph, S. & Goswami, B. N. A self-organizing map-based ensemble forecast system for extended range prediction of active/break cycles of Indian summer monsoon. *Journal of Geophysical Research (Atmospheres)* **118**, 9022–9034 (2013).

49. Giffard-Roisin, S. *et al.* Tropical Cyclone Track Forecasting Using Fused Deep Learning From Aligned Reanalysis Data. *Frontiers in Big Data* **3**, 1 (2020).
50. Lorenz, E. N. Deterministic Nonperiodic Flow. *Journal of Atmospheric Sciences* **20**, 130–141 (1963).
51. Chattopadhyay, R. *et al.* Large-scale teleconnection patterns of Indian summer monsoon as revealed by CFSv2 retrospective seasonal forecast runs. *International Journal of Climatology* **36**, 3297–3313 (2016).
52. Hoskins, B. The potential for skill across the range of the seamless weather-climate prediction problem: a stimulus for our science. *Quarterly Journal of the Royal Meteorological Society* **139**, 573–584 (2013).
53. Saha, M., Santara, A., Mitra, P., Chakraborty, A. & Nanjundiah, R. S. Prediction of the Indian summer monsoon using a stacked autoencoder and ensemble regression model. *International Journal of Forecasting* **37**, 58–71 (2021).
54. Ham, Y.-G., Kim, J.-H. & Luo, J.-J. Deep learning for multi-year ENSO forecasts. *Nature* **573**, 568–572 (2019).
55. Nooteboom, P. D., Feng, Q. Y., López, C., Hernández-García, E. & Dijkstra, H. A. Using network theory and machine learning to predict El Niño. *Earth Syst. Dynam.* **9**, 969–983 (2018).
56. Sikka, D. R. Some aspects of the large scale fluctuations of summer monsoon rainfall over India in relation to fluctuations in the planetary and regional scale circulation parameters. *Proceedings of the Indian Academy of Sciences - Earth and Planetary Sciences* **89**, 179–195 (1980).
57. Ashok, K., Behera, S. K., Rao, S. A., Weng, H. & Yamagata, T. El Niño Modoki and its possible teleconnection. *Journal of Geophysical Research: Oceans* **112**, (2007).
58. Ashok, K., Guan, Z., Saji, N. H. & Yamagata, T. Individual and Combined Influences of ENSO and the Indian Ocean Dipole on the Indian Summer Monsoon. *Journal of Climate* **17**, 3141–3155 (2004).
59. Goswami, B. N., Venugopal, V., Sengupta, D., Madhusoodanan, M. S. & Xavier, P. K. Increasing Trend of Extreme Rain Events Over India in a Warming Environment. *Science* **314**, 1442 (2006).
60. Krishnan, R. & Sugi, M. Pacific decadal oscillation and variability of the Indian summer monsoon rainfall. *Climate Dynamics* **21**, 233–242 (2003).
61. Singh, M. *et al.* Fingerprint of volcanic forcing on the ENSO–Indian monsoon coupling. *Sci Adv* **6**, eaba8164 (2020).
62. Ayantika, D. C. *et al.* Understanding the combined effects of global warming and anthropogenic aerosol forcing on the South Asian monsoon. *Climate Dynamics* **56**, 1643–1662 (2021).
63. Fadnavis, S. *et al.* Atmospheric Aerosols and Trace Gases. in *Assessment of Climate Change over the Indian Region: A Report of the Ministry of Earth Sciences (MoES), Government of India* (eds. Krishnan, R. et al.) 93–116 (Springer Singapore, 2020). doi:10.1007/978-981-15-4327-2_5.
64. de Witt, C. S. & Hornigold, T. Stratospheric Aerosol Injection as a Deep Reinforcement Learning Problem. *arXiv.org* (2019) doi:arXiv:1905.07366.
65. Seifert, A. & Rasp, S. Potential and Limitations of Machine Learning for Modeling Warm-Rain Cloud Microphysical Processes. *Journal of Advances in Modeling Earth Systems* **12**, e2020MS002301 (2020).
66. Singh, B. B. *et al.* Linkage of water vapor distribution in the lower stratosphere to organized Asian summer monsoon convection. *Climate Dynamics* (2021) doi:10.1007/s00382-021-05772-2.
67. Geer, A. J. Learning earth system models from observations: machine learning or data assimilation? *Philosophical Transactions of the Royal Society A: Mathematical, Physical and Engineering Sciences* **379**, 20200089 (2021).
68. Grönquist, P. *et al.* Deep learning for post-processing ensemble weather forecasts. *Philosophical Transactions of the Royal Society A: Mathematical, Physical and Engineering Sciences* **379**, 20200092 (2021).
69. Kashinath, K. *et al.* Physics-informed machine learning: case studies for weather and climate modelling. *Philosophical Transactions of the Royal Society A: Mathematical, Physical and Engineering Sciences* **379**, 20200093 (2021).
70. Balaji, V. Climbing down Charney's ladder: machine learning and the post-Dennard era of computational climate science. *Philosophical Transactions of the Royal Society A: Mathematical, Physical and Engineering Sciences* **379**, 20200085 (2021).
71. Pulkkinen, S. *et al.* Pysteps: an open-source Python library for probabilistic precipitation nowcasting (v1.0). *Geosci. Model Dev.* **12**, 4185–4219 (2019).
72. Kim, T.-J. & Kwon, H.-H. Development of Tracking Technique for the Short Term Rainfall Field Forecasting. *Procedia Engineering* **154**, 1058–1063 (2016).
73. Agrawal, S. *et al.* Machine Learning for Precipitation Nowcasting from Radar Images. *arXiv.org* (2019) doi:https://arxiv.org/abs/1912.12132.
74. Su, A., Li, H., Cui, L. & Chen, Y. A Convection Nowcasting Method Based on Machine Learning. *Advances in Meteorology* **2020**, 5124274 (2020).

75. Arulraj, M. & Barros, A. P. Automatic detection and classification of low-level orographic precipitation processes from space-borne radars using machine learning. *Remote Sensing of Environment* **257**, 112355 (2021).
76. Sarafanov, M., Kazakov, E., Nikolay, N. O. & Kalyuzhnaya, A. V. A Machine Learning Approach for Remote Sensing Data Gap-Filling with Open-Source Implementation: An Example Regarding Land Surface Temperature, Surface Albedo and NDVI. **12**, 3865 (2020).
77. R. Cresson, D. Ienco, R. Gaetano, K. Ose, & D. H. Tong Minh. Optical image gap filling using deep convolutional autoencoder from optical and radar images. in *IGARSS 2019 - 2019 IEEE International Geoscience and Remote Sensing Symposium* 218–221 (2019). doi:10.1109/IGARSS.2019.8900353.
78. Boukabara, S.-A. *et al.* Leveraging Modern Artificial Intelligence for Remote Sensing and NWP: Benefits and Challenges. *Bulletin of the American Meteorological Society* **100**, ES473–ES491 (2019).